\documentclass[sigconf]{acmart}
\AtBeginDocument{%
  }

\setcopyright{acmlicensed}

\copyrightyear{2026}
\acmYear{2026}
\setcopyright{cc}
\setcctype{by-nc-nd}
\acmConference[UbiComp Companion '26]{Companion of the 2026 ACM International Joint Conference on Pervasive and Ubiquitous Computing}{October 11--15, 2026}{Shanghai, China}
\acmBooktitle{Companion of the 2026 ACM International Joint Conference on Pervasive and Ubiquitous Computing (UbiComp Companion '26), October 11--15, 2026, Shanghai, China}
\acmDOI{10.1145/3798063.3837408}
\acmISBN{979-8-4007-2533-3/2026/10}

\usepackage{booktabs}
\usepackage[table,xcdraw]{xcolor}
\usepackage{balance}

\begin{document}

\title{\textit{Position: Robot Privacy as Embodied Boundary Work} \\
Connecting Capabilities, Contexts, and Design Responses in Everyday Robotics
}

\author{Liwen He}
\orcid{0000-0003-0715-9252}
\affiliation{%
    \institution{The Hong Kong University of Science and Technology (Guangzhou)}
    \city{Guangzhou}
    \state{Guangdong}
    \country{China}
    \institution{Tsinghua University}
    \city{Beijing}
    \country{China}}
\email{helw24@mails.tsinghua.edu.cn}

\author{Shuning Zhang}
\orcid{0000-0002-4145-117X}
\affiliation{%
    \institution{Tsinghua University}
    \city{Beijing}
    \country{China}}
\email{zsn23@mails.tsinghua.edu.cn}

\author{Chengwen Zhang}
\orcid{0009-0003-4285-7192}
\affiliation{%
    \institution{Tsinghua University}
    \city{Beijing}
    \country{China}}
\email{zcw25@mails.tsinghua.edu.cn}

\author{Xin Yi}
\orcid{0000-0001-8041-7962}
\affiliation{%
    \institution{Tsinghua University}
    \institution{Beijing Academy of Artificial Intelligence}
    \city{Beijing}
    \country{China}}
\email{yixin@tsinghua.edu.cn}

\author{Chun Yu}
\orcid{0000-0003-2591-7993}
\affiliation{%
    \institution{Tsinghua University}
    \city{Beijing}
    \country{China}}
\email{chunyu@tsinghua.edu.cn}

\author{Jihong Jeung}
\orcid{0009-0005-8925-0712}
\affiliation{%
    \institution{Tsinghua University}
    \city{Beijing}
    \country{China}}
\email{jihong95@mail.tsinghua.edu.cn}

\author{Xin Tong}
\authornote{Corresponding author.}
\orcid{0000-0002-8037-6301}
\affiliation{%
    \institution{The Hong Kong University of Science and Technology (Guangzhou)}
    \city{Guangzhou}
    \state{Guangdong}
    \country{China}
    \department{}
    \institution{The Hong Kong University of Science and Technology}
    \city{Hongkong}
    \country{China}}
\email{xint@hkust-gz.edu.cn}

\renewcommand{\shortauthors}{Liwen He, Shuning Zhang, Chengwen Zhang, Xin Yi, Chun Yu, Jihong Jeung, Xin Tong}

\begin{abstract}
  Robots are increasingly entering everyday environments where privacy is shaped not only by data practices, but also by spatial, bodily, social, and relational boundaries. Their embodied capabilities allow them to reshape these boundaries through situated action, challenging privacy framings centered on data flows, interface settings, or one-time consent. Prior work has examined robot privacy through sensing, data collection, telepresence, transparency, consent, bystander awareness, and multi-stakeholder governance. Building on this work, we propose embodied boundary privacy as a \textbf{capability-by-context framing} for examining how physically present robots may reshape privacy boundaries in situated interaction. Specifically, this framing organizes privacy risks across seven robot capabilities and five deployment contexts, asking how embodied capabilities enable boundary crossings and how situated contexts shape who is affected, how these crossings are interpreted, and when they become contested. We use this perspective to outline design and research implications for embodied privacy mechanisms, including boundary checkpoints, viewpoint-aware sensing control, remote-presence disclosure, object- and body-level access rules, constraints on socially persuasive privacy influence, and local interruption rights. We encourage HRI research, design, and governance to treat robot movement, orientation, proximity, object access, remote presence, and social expression as privacy-relevant actions whose meaning depends on context.
\end{abstract}





\maketitle

\section{Introduction}
Robots are increasingly entering everyday environments where privacy is not only managed through data settings, but also through spatial, bodily, and social boundaries \cite{cardiell2021robot}. Domestic robots move between rooms \cite{stapels2023never}, care robots approach bodies and support intimate routines \cite{grabler2025privacy}, telepresence robots allow remote people to appear in local spaces \cite{hubers2015video}, and public service robots observe and navigate around bystanders \cite{aryania2026impact}. In these contexts, privacy is experienced through questions such as who can enter a room, who can come close, who can observe an activity, who can access an object, and who can participate in an interaction. These questions are not only informational. They are also embodied, situated, and relational \cite{levinson2024snitches}.

Prior work has made substantial progress in robot privacy by examining sensing, data collection, cloud processing, transparency, consent, access control, telepresence, bystander awareness, and multi-stakeholder governance \cite{horstmann2020towards}. Recent work has further argued that robot privacy extends beyond data protection to include physical, bodily, psychological, and relational dimensions \cite{horstmann2020towards,hofmann2013ethical}, and has discussed capabilities such as mobility, teleoperation, manipulation, and social presence as sources of privacy risk \cite{chen2025exploring}. However, existing framings still tend to organize robot privacy around data practices, deployment domains, or broad categories of harm \cite{levinson2025effective}. What remains less explicit is how embodiment changes the conditions of privacy boundary access. Robots do not only collect information; they can enter spaces, change viewpoints, approach bodies, access objects, mediate remote presence, persist in sensitive situations, and use social expression to make access feel acceptable or difficult to refuse \cite{lutz2024social}. This shifts the central question from only “\textit{what data is collected}” to also “\textit{what forms of access a robot can enact, how, where, for whom, and under whose control.}”

\begin{figure}
    \centering
    \includegraphics[width=1\linewidth]{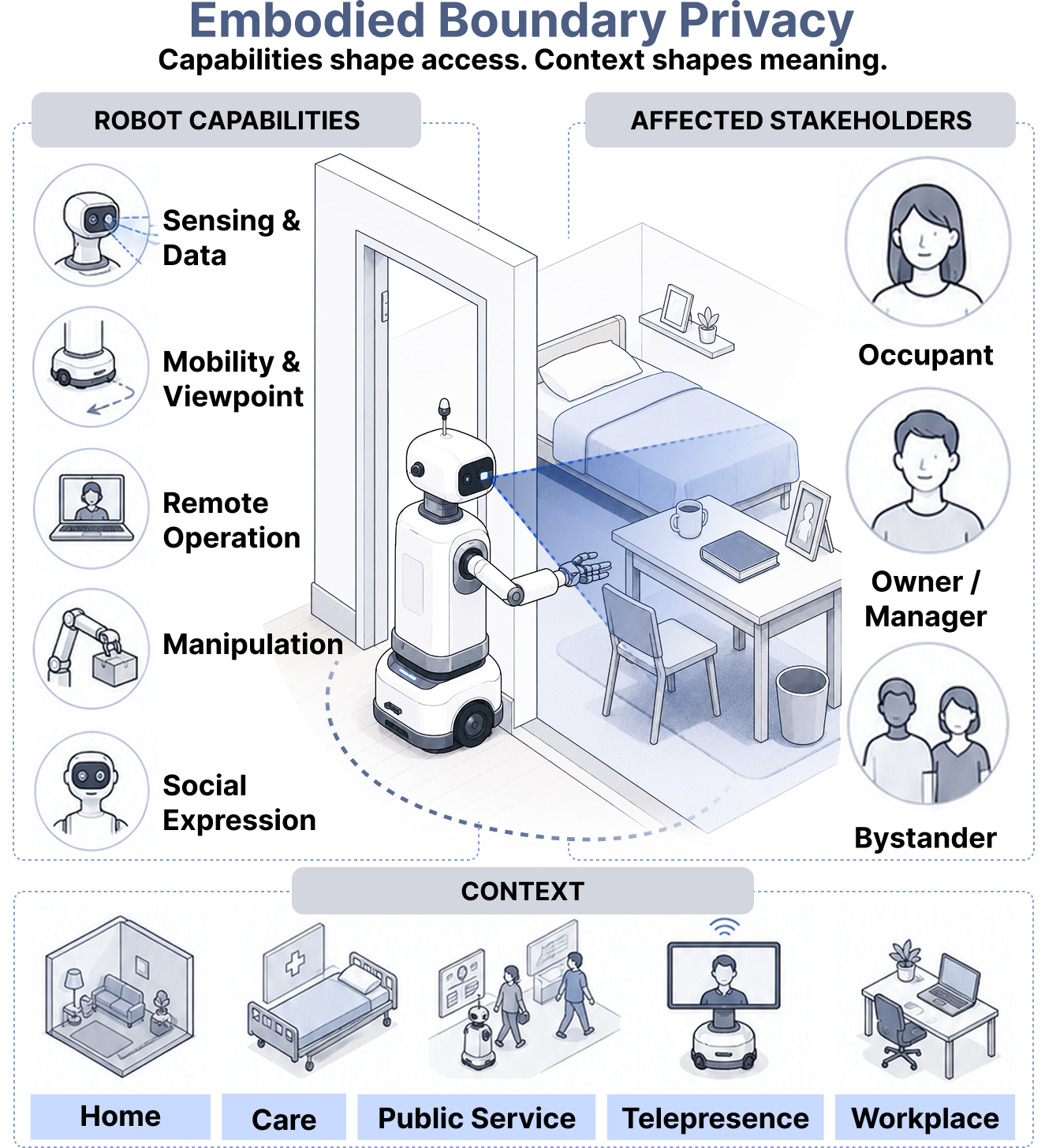}
    \caption{Overview of the embodied boundary privacy framework.}
    \Description{Conceptual illustration of the embodied boundary privacy framework. 
At the center, a mobile service robot crosses a doorway into a private room while its sensing field extends into the space, illustrating embodied access across a spatial boundary. 
Five robot capability groups are shown on the left: sensing and data, mobility and viewpoint, remote operation, manipulation, and social expression. 
Affected stakeholders are shown on the right, including the occupant, owner or manager, and bystander. 
Five representative deployment contexts are shown at the bottom: home, care, public service, telepresence, and workplace. 
The figure illustrates that robot capabilities shape how access occurs, while context and stakeholder relations shape the privacy meaning of that access.
}
    \label{fig:framework}
\end{figure}

Building on this gap, we propose embodied boundary privacy as a capability-by-context framing for understanding privacy in physically situated human-robot interaction (see Figure \ref{fig:framework}). This framing treats robot privacy risks as emerging from the interaction between what a robot is capable of doing and where, with whom, and under what social conditions those capabilities are enacted. A capability such as mobility, viewpoint control, manipulation, or social expression does not have a fixed privacy meaning on its own. Its implications depend on whether the robot is crossing a bedroom threshold, approaching a care recipient’s body, observing bystanders in a public setting, mediating a remote operator, or presenting itself as a friendly companion. In this sense, robots do not simply collect information within a context; they actively reorganize access, attention, intimacy, and control through embodied action.


To describe these robot-specific privacy concerns, we use embodied boundary privacy to refer to people’s ability to perceive, negotiate, and maintain privacy boundaries around information, spaces, bodies, objects, relationships, and decisions when interacting with physically present robots \cite{grasso2025designing,windl2024privacy}. This perspective complements existing work on data protection, transparency, and consent by shifting attention from isolated data flows or one-time permission decisions to the ways privacy boundaries are enacted over the course of interaction. It asks not only what data a robot collects, but also where it can go, how close it can get, what it can access, who can observe through it, how it signals its intentions, and how affected people can accept, refuse, interrupt, or renegotiate its presence.


This position paper makes three contributions. First, we articulate embodied boundary privacy as a robot-specific privacy framing that explains how physically present robots reshape boundaries around spaces, bodies, objects, relationships, information, and decisions. Second, we develop a capability-by-context account of robot privacy risk, showing how seven robot capability categories create different pathways of boundary crossing across everyday contexts. Third, we translate this framing into a design and research agenda for embodied privacy mechanisms, including boundary checkpoints, viewpoint-aware sensing control, remote-presence disclosure, object- and body-level access rules, limits on socially persuasive privacy influence, and local interruption rights for affected users and bystanders.

\section{Existing Framings of Robot Privacy}
Prior work has made substantial progress in identifying privacy risks in non-industrial robots. We organize this work around three related framings: robots as sensing and data-processing systems, robots as remotely accessible and networked systems, and robots as social actors in shared contexts. These framings provide the basis for our position. Together, they show that robot privacy is shaped by data practices, mediated access, social relationships, deployment settings, and affected stakeholders. Building on these strands, we examine how robot capabilities become privacy-relevant when they are enacted through movement, viewpoint, proximity, object access, remote presence, and social expression.

\subsection{Robots as sensing and data-processing systems}
A prominent line of HRI and robot-privacy research frames non-industrial robots as networked sensing and data-processing systems. From this perspective, privacy concerns arise across the data lifecycle, including collection, inference, storage, access control, disclosure, and secondary use \cite{lutz2019privacy,horstmann2020towards,sullivan2025protecting,sedenberg2016designing,callander2024navigating}. This framing is well motivated. Social and service robots often rely on cameras, microphones, depth sensors, localization and mapping, cloud services, and other forms of context awareness to navigate, personalize interaction, and support users \cite{lutz2019privacy,heuer2021privacy,schulz2018privacy,sullivan2025benchmarking}. Foundational work also shows that household robots extend familiar IoT privacy concerns by combining persistent sensing with mobility and remote connectivity \cite{denning2009spotlight}.

This literature has identified privacy risks that go beyond raw sensor capture. Robot sensing can reveal routines, occupancy patterns, conversations, social relationships, health-related attributes, affective states, and longitudinal profiles, especially in domestic, care, and public or semi-public settings involving vulnerable users or multiple stakeholders \cite{grabler2025privacy,fronemann2022should,bhat2022confused,zhong2025values}. It has also informed privacy-by-design strategies such as data minimization, careful sensor selection, lower-fidelity sensing where appropriate, local or on-device processing, reduced cloud dependency, inspection and deletion interfaces, and configurable privacy settings \cite{heuer2021privacy,eick2020enhancing,sedenberg2016designing,horstmann2020towards,sullivan2025protecting}. Work on transparency further emphasizes explanations of what data are collected and why, as well as embodied indicators of sensing, recording, or deactivation states \cite{grasso2025designing,aryania2026impact,grasso2024investigating,shinohara2023understanding,nieto2025robot}.

This body of work establishes sensing, inference, and data governance as central concerns for robot privacy. It also provides a foundation for examining how sensing is situated in embodied interaction. For physically present robots, privacy-relevant sensing is often tied to where the robot moves, which direction it faces, how close it comes, what activity it approaches, and whether nearby people can notice or control that sensing. We build on data-centered robot privacy work by considering how sensing and inference are coupled with embodied access in everyday environments.

\begin{table*}[]
\caption{Embodied Boundary Privacy Framework}
\label{tab:framework}
\begin{tabular}{@{}llllll@{}}
\toprule
\textbf{Context} & \textbf{Sensing \& data} & \textbf{Mobility \& viewpoint} & \textbf{Remote operation} & \textbf{Manipulation} & \textbf{Social expression} \\ \midrule
\rowcolor[HTML]{EFEFEF} 
\textbf{Home} & \begin{tabular}[c]{@{}l@{}}Recording, mapping, \\ routine inference\end{tabular} & \begin{tabular}[c]{@{}l@{}}Access to rooms \\ and thresholds\end{tabular} & \begin{tabular}[c]{@{}l@{}}Remote access to \\ family spaces\end{tabular} & \begin{tabular}[c]{@{}l@{}}Exposure of \\ personal objects\end{tabular} & \begin{tabular}[c]{@{}l@{}}Attachment, \\ over-disclosure\end{tabular} \\
\textbf{Care} & \begin{tabular}[c]{@{}l@{}}Health/emotion \\ inference\end{tabular} & \begin{tabular}[c]{@{}l@{}}Night patrols and \\ room checks\end{tabular} & Clinical check-ins & \begin{tabular}[c]{@{}l@{}}Bathing, toileting, \\ dressing privacy\end{tabular} & \begin{tabular}[c]{@{}l@{}}Dignity, comfort, \\ feeling judged\end{tabular} \\
\rowcolor[HTML]{EFEFEF} 
\textbf{Telepresence} & \begin{tabular}[c]{@{}l@{}}Platform access \\ leakage\end{tabular} & \begin{tabular}[c]{@{}l@{}}Mobile observation\\ of local spaces\end{tabular} & \begin{tabular}[c]{@{}l@{}}Operator identity \\ and framing\end{tabular} & — & \begin{tabular}[c]{@{}l@{}}Intensified perceived \\ presence\end{tabular} \\
\textbf{Public service} & \begin{tabular}[c]{@{}l@{}}Service data \\ collection\end{tabular} & \begin{tabular}[c]{@{}l@{}}crowd video and \\ navigation traces\end{tabular} & \begin{tabular}[c]{@{}l@{}}Remote service \\ operation\end{tabular} & \begin{tabular}[c]{@{}l@{}}Transaction/object \\ logs\end{tabular} & \begin{tabular}[c]{@{}l@{}}Friendliness lowers \\ vigilance\end{tabular} \\
\rowcolor[HTML]{EFEFEF} 
\textbf{\begin{tabular}[c]{@{}l@{}}Accessibility / \\ bystanders\end{tabular}} & \begin{tabular}[c]{@{}l@{}}Hard-to-notice \\ capture\end{tabular} & \begin{tabular}[c]{@{}l@{}}Unclear camera \\ direction\end{tabular} & \begin{tabular}[c]{@{}l@{}}Invisible operator \\ identity\end{tabular} & — & \begin{tabular}[c]{@{}l@{}}Unequal perception \\ of gaze\end{tabular} \\ \bottomrule
\end{tabular}
\end{table*}

\subsection{Robots as remotely accessible and networked systems}

A second framing views robots as networked systems that mediate access between local environments and remote actors. This is most visible in telepresence and avatar robots, where a mobile platform enables a remote operator to move through a home, office, care facility, or other semi-public settings. In such configurations, privacy risk is not only a matter of downstream data storage or sharing, but also concerns who is effectively “present” through the robot, what they can see/hear, how freely they can navigate, and whether local people and bystanders can recognize, understand, or restrict this access \cite{parts2025systematic,butler2015privacy,krupp2017focus}. Early security-oriented work already highlighted that networked household robots can become remote “sensing and actuation” endpoints, enabling surveillance-like access if compromised or misconfigured, thereby coupling privacy concerns with remote connectivity and control \cite{denning2009spotlight}.

Telepresence research has also produced some of the most concrete robot-specific privacy interventions, especially for visual/observational privacy. A well-developed line of work formalizes observational privacy goals (e.g., preventing remote operators from identifying or discerning sensitive details) and evaluates video manipulation techniques such as blurring, redaction, replacement, and abstraction to protect local spaces and people while preserving remote navigation or task utility \cite{hubers2015video,klow2017privacy}. Complementing these sensing-side interventions, research on telepresence privacy perceptions shows that privacy judgments are strongly shaped by social framing and operator identity (e.g., family member vs stranger), and that such framing effects can persist across scenarios even when the technical capabilities remain unchanged  \cite{rueben2017framing}. Qualitative work (e.g., focus groups) further documents local users’ privacy concerns around what remote operators can observe and how much control they should have, reinforcing that telepresence privacy is simultaneously technical (sensor access) and social (relationship and legitimacy of access) \cite{krupp2017focus}. Extending the threat model, recent work on avatar robots explicitly considers malicious or low-moral teleoperators, mapping how mobility and onboard cameras/microphones can be used for privacy violations and related harms \cite{shaheen2024investigation}.

This framing is useful for showing that robot privacy involves mediated presence and distributed access authority. The people physically affected by a robot may not be the same people who own, configure, operate, or benefit from it. Cloud-connected domestic, care, and service robots further complicate this situation because vendors, developers, caregivers, employers, relatives, or third-party services may gain access to robot-generated data or capabilities \cite{reinhardt2025toward,horstmann2020towards,grabler2025privacy,levinson2024snitches}. We take this insight as a starting point for considering how remote access is experienced locally through the robot’s movement, viewpoint, proximity, operator disclosure, and available override mechanisms.

\subsection{Robots as social actors in shared contexts}

A third framing treats robot privacy as a social and contextual issue rather than only a matter of individual data control. Social robots are often designed with expressive movements, affective behaviors, and socially legible roles that shape how people interpret and engage with them \cite{he2026pets}. Prior work emphasizes that non-industrial robots are often deployed in multi-person settings with asymmetric relationships and sensitive social roles, where one robot’s sensing, movement, and interaction behavior may affect several stakeholders at once \cite{fronemann2022should}. Domestic robots may operate around family members, children, adolescents, visitors, and older adults, raising questions about shared ownership, family boundary management, and children’s autonomy \cite{levinson2024snitches}. Care robots may involve patients, caregivers, relatives, clinicians, and institutions, creating tensions among safety, monitoring, autonomy, dignity, and bodily privacy \cite{zhong2025values}. Public and semi-public service robots may affect customers, workers, passersby, and bystanders who never explicitly opt into the interaction \cite{hedaoo2019robot,bhat2022confused}.

This work shows that privacy expectations vary by relationship, role, location, urgency, vulnerability, and power. It has drawn attention to bystander privacy, multi-user privacy, child and adolescent privacy, elder privacy, associational privacy, relational privacy, and contextually appropriate disclosure \cite{tang2022confidant,jayaraman2024social}. It also motivates designs such as bystander-facing transparency, accessible cues, household or group-level configuration, context-aware disclosure rules, privacy conversations, and multi-touchpoint support \cite{grasso2024investigating,grasso2025designing}.

Social and contextual framings are valuable because they move robot privacy beyond one-time consent and static settings, and instead characterize privacy as negotiated among stakeholders with different roles, vulnerabilities, and power \cite{levinson2024snitches}. They also suggest privacy designs that support (i) bystander-facing transparency and accessible cues, (ii) household- or group-level configuration, (iii) context-aware disclosure and interaction rules, and (iv) ongoing negotiation mechanisms such as privacy conversations and multi-touchpoint support, including escalation to human support when needed \cite{grasso2024investigating}.We extend this contextual insight by asking how different robot capabilities materialize privacy expectations as concrete boundary issues in interaction. For example, the privacy implications of a mobile camera, a remote clinical check-in, a care robot approaching a body, or a companion-like robot inviting disclosure depend on both the capability involved and the context in which it is enacted. This motivates a perspective that connects robot capabilities, deployment contexts, and the boundaries that people need to perceive, negotiate, and maintain.

Taken together, these three framings show that robot privacy is shaped by data practices, mediated access, and situated social relationships. Our position builds on this foundation by foregrounding embodied capability: how robots sense, move, orient, approach, manipulate, express social roles, and connect local environments to remote actors or institutions. This perspective does not replace existing framings. Rather, it organizes them around the boundary work that robots perform in everyday environments.

\section{Embodied Boundary Privacy Framework}

\begin{table*}[]
\caption{Embodied Robot Capabilities and Privacy Boundary Mechanisms}
\label{tab:capability}
\begin{tabular}{@{}llll@{}}
\toprule
\multicolumn{1}{c}{\textbf{Embodied capability}} & \multicolumn{1}{c}{\textbf{Concrete capability}} & \multicolumn{1}{c}{\textbf{Boundary affected}} & \multicolumn{1}{c}{\textbf{Privacy risk mechanism}} \\ \midrule
\rowcolor[HTML]{EFEFEF} 
\textbf{Mobility} & \begin{tabular}[c]{@{}l@{}}Wheeled or \\ aerial movement\end{tabular} & \begin{tabular}[c]{@{}l@{}}Spatial, \\ observational, \\ bystander\end{tabular} & \begin{tabular}[c]{@{}l@{}}Crossing thresholds, \\ entering private zones, \\ enabling mobile observation\end{tabular} \\
\textbf{\begin{tabular}[c]{@{}l@{}}Vision and \\ spatial sensing\end{tabular}} & \begin{tabular}[c]{@{}l@{}}RGB, depth, SLAM,\\ thermal cameras\end{tabular} & \begin{tabular}[c]{@{}l@{}}Informational, \\ spatial\end{tabular} & \begin{tabular}[c]{@{}l@{}}Capturing faces, \\ reconstructing rooms, \\ inferring routines and \\ household layout\end{tabular} \\
\rowcolor[HTML]{EFEFEF} 
\textbf{Manipulation} & \begin{tabular}[c]{@{}l@{}}Arms, grippers, \\ door/object handling\end{tabular} & \begin{tabular}[c]{@{}l@{}}Bodily, \\ territorial, \\ object privacy\end{tabular} & \begin{tabular}[c]{@{}l@{}}Opening drawers or doors, \\ exposing personal objects, \\ touching or assisting bodies\end{tabular} \\
\textbf{\begin{tabular}[c]{@{}l@{}}Physiological and \\ affective sensing\end{tabular}} & \begin{tabular}[c]{@{}l@{}}HRV, EDA, posture, \\ facial expression\end{tabular} & \begin{tabular}[c]{@{}l@{}}Informational, \\ psychological\end{tabular} & \begin{tabular}[c]{@{}l@{}}Inferring emotion, stress, \\ health status, or vulnerability\end{tabular} \\
\rowcolor[HTML]{EFEFEF} 
\textbf{Social expression} & \begin{tabular}[c]{@{}l@{}}Eyes, gaze, facial display, \\ voice, pet-like behavior\end{tabular} & \begin{tabular}[c]{@{}l@{}}Psychological, \\ relational, \\ decisional\end{tabular} & \begin{tabular}[c]{@{}l@{}}Increasing trust, \\ perceived judgment, \\ attachment, or over-disclosure\end{tabular} \\
\textbf{\begin{tabular}[c]{@{}l@{}}Remote operation and \\ multi-party access\end{tabular}} & \begin{tabular}[c]{@{}l@{}}Teleoperation, cloud control,\\ family/institutional access\end{tabular} & \begin{tabular}[c]{@{}l@{}}Relational, \\ decisional, \\ accountability\end{tabular} & \begin{tabular}[c]{@{}l@{}}Shifting observation rights, \\ weakening consent, \\ creating access conflicts\end{tabular} \\
\rowcolor[HTML]{EFEFEF} 
\textbf{\begin{tabular}[c]{@{}l@{}}Networking and \\ cloud analysis\end{tabular}} & \begin{tabular}[c]{@{}l@{}}Logs, NLP inference, \\ platform integration\end{tabular} & \begin{tabular}[c]{@{}l@{}}Informational, \\ infrastructural\end{tabular} & \begin{tabular}[c]{@{}l@{}}Long-term profiling, \\ secondary use, \\ opaque data flows\end{tabular} \\ \bottomrule
\end{tabular}
\end{table*}

Our aim is to use embodiment to explain the mechanisms through which privacy boundaries are crossed. A robot may violate privacy not only by storing video, but by moving a camera across a threshold, orienting its sensors toward a private activity, approaching a body, opening a container, or presenting itself as a trusted companion during disclosure. In this sense, embodiment is not merely the object of privacy concern; it is the means through which privacy boundaries become crossed, blurred, or renegotiated.

Embodied boundary privacy is organized around three analytical questions:
(1) What capability enables access?
(2) What boundary is crossed or reshaped?
(3) Who can perceive, negotiate, refuse, or repair that crossing in context?

Embodied boundary privacy complements relational privacy, bystander privacy, and other contextual accounts rather than replacing them. While these perspectives help explain whose privacy interests and expectations matter, our framing focuses specifically on how robot capabilities enact or reshape access to spaces, bodies, objects, information, and social interactions. Its core scope is therefore privacy change produced through physically situated robotic action, rather than downstream data use that is independent of such embodied access.

\subsection{From privacy dimensions to boundary mechanisms}
We propose a capability × context view to examine how robot privacy risks are produced in everyday environments (see Table \ref{tab:framework}). The privacy implications of a ``home robot,'' ``care robot,'' or ``public service robot'' depend, in part, on the capabilities it brings into a specific setting: whether it only listens, moves through rooms, changes its viewpoint, supports remote operation, touches bodies or objects, appears in human-like or pet-like forms, or connects its sensing system to broader data infrastructures. 
The same capability may carry different privacy meanings across contexts: mobility may involve crossing room thresholds at home, conducting room checks in care, or navigating among bystanders in public settings; manipulation may expose personal objects at home or support intimate bodily care; and remote operation may enable a clinical check-in or provide access to family spaces. These contrasts show that capabilities shape the mode of boundary crossing, while context determines the affected stakeholders, expected boundaries, and authority relations.
Table~\ref{tab:capability} details seven capability categories, while Table~\ref{tab:framework} groups related capabilities into five broader families to illustrate their intersections with deployment contexts.

The matrix in Table~\ref{tab:framework} maps representative intersections between robot capabilities and deployment contexts. The horizontal axis identifies capabilities that shape the mode of boundary crossing. The vertical axis identifies contexts that shape the affected stakeholders, authority relations, and conditions under which boundary crossing is permitted, normalized, or contested. Together, the two dimensions show how robot privacy risks arise through the coupling of embodied capability and situated social context: capabilities define possible forms of access, while contexts define their social consequences.

\subsection{Capability explains the mechanism of boundary crossing}
The capability axis captures the robot-side mechanism of privacy risk. Stationary sensing primarily extends familiar IoT concerns: audio, video, and interaction traces can be captured over time. Mobile vision changes the problem by allowing the robot to carry sensing across spatial boundaries. It can enter rooms, follow users, approach activities, and produce maps or routines from movement. Remote operation introduces a different mechanism: the robot becomes a proxy through which another person can observe, navigate, or act in a local environment. Manipulation expands privacy from observation to access, because the robot can touch bodies, open containers, move personal objects, or expose things that were previously protected by physical separation.

Social expression adds another layer. Human-like and pet-like robots may not collect different data from less expressive robots, but they can change how people interpret the interaction. They may invite trust, attachment, compliance, or disclosure, making privacy risk partly psychological and relational. Aerial mobility further destabilizes familiar spatial boundaries by introducing viewpoints that bypass doors, walls, windows, and bodily distance. Multimodal networking connects local sensing to broader infrastructures of inference, storage, remote access, and institutional accountability. These examples show why embodiment matters: privacy risk is not only attached to the sensor, but to the sensor’s mobility, orientation, social framing, access to objects, and connection to other actors.

\subsection{Context explains the social distribution of risk}
The context axis captures how the same capability becomes meaningful in a particular social setting. In the home, privacy is distributed across household members, children, visitors, and remote family members. A robot’s data and movement may be configured by one person while affecting others. In care settings, boundary crossing is often justified through safety, assistance, or clinical responsibility, but the same justification may weaken bodily privacy, dignity, and autonomy. In telepresence, the central issue is not only what the robot can sense, but who becomes present through it, how that operator is framed, and whether local people can recognize or limit that presence.

In public service settings, robot privacy extends beyond the direct user to customers, workers, passersby, and bystanders who may never explicitly interact with the robot. In accessibility and bystander situations, the problem is not only exposure but perceptibility: some people may not be able to identify where the robot is looking, whether it is recording, who is operating it, or how its data flows. Context therefore determines whether a capability is experienced as assistance, companionship, care, convenience, surveillance, or control. It also determines who can consent, who can refuse, and who is merely made visible.

This capability × context view reframes robot privacy from a question of ``\textit{what data does the robot collect?}'' to a more situated question: ``\textit{what boundary does this capability cross in this context, and who has the power to negotiate it?}'' This reframing prepares the design implications that follow. If privacy risk is produced through embodied boundary crossing, then privacy design must address not only data transparency and settings, but also movement, proximity, access, social cues, remote presence, and multi-party negotiation.

\begin{table*}[]
\caption{Representative Deployment Contexts and Privacy Concern Clusters in Robot Privacy}
\label{tab:context-robot-privacy}
\begin{tabular}{@{}llll@{}}
\toprule
\textbf{Context / Robot domain} & \textbf{Deployment contexts} & \textbf{Primary stakeholders} & \textbf{Privacy concern clusters} \\ \midrule
\rowcolor[HTML]{EFEFEF} 
\textbf{\begin{tabular}[c]{@{}l@{}}\textit{Home}: \\ domestic, social, and \\ companion robots\end{tabular}} & \begin{tabular}[c]{@{}l@{}}Private homes,\\ including bedrooms, \\ living rooms, kitchens, \\ homework spaces, and \\ therapy areas\end{tabular} & \begin{tabular}[c]{@{}l@{}}Adults, children, \\ teenagers, families, \\ older adults, \\ therapists\end{tabular} & \begin{tabular}[c]{@{}l@{}}Continuous multimodal sensing; \\ home mapping and routine inference; \\ family boundary management; \\ children’s data co-ownership; \\ emotional dependency and \\ intimate disclosure\end{tabular} \\
\textbf{\begin{tabular}[c]{@{}l@{}}\textit{Car}e: \\ socially assistive robots\\ and elder-care systems\end{tabular}} & \begin{tabular}[c]{@{}l@{}}Nursing homes, \\ assisted living facilities, \\ and private residences \\ with care services\end{tabular} & \begin{tabular}[c]{@{}l@{}}Care recipients, \\ caregivers, relatives, \\ clinicians, \\ care institutions\end{tabular} & \begin{tabular}[c]{@{}l@{}}Bodily privacy in intimate care; \\ safety–autonomy–privacy tensions; \\ dignity and calmness; \\ multi-stakeholder data access conflicts\end{tabular} \\
\rowcolor[HTML]{EFEFEF} 
\textbf{\begin{tabular}[c]{@{}l@{}}\textit{Telepresence}: \\ telepresence and \\ avatar robots\end{tabular}} & \begin{tabular}[c]{@{}l@{}}Homes, offices, clinics, \\ care institutions, and \\ semi-public buildings\end{tabular} & \begin{tabular}[c]{@{}l@{}}Remote operators, \\ local users, hosts, \\ bystanders\end{tabular} & \begin{tabular}[c]{@{}l@{}}Visual and observational privacy; \\ viewpoint control; operator identity ambiguity; \\ unintended room access; platform compromise\end{tabular} \\
\textbf{\begin{tabular}[c]{@{}l@{}}\textit{Public and semi-public}\\ \textit{service}: service robots\end{tabular}} & \begin{tabular}[c]{@{}l@{}}Cafés, hotels, malls, \\ airports, offices, and \\ retail stores\end{tabular} & \begin{tabular}[c]{@{}l@{}}Customers, employees, \\ passersby, \\ service providers\end{tabular} & \begin{tabular}[c]{@{}l@{}}Short-term personal data disclosure; \\ ambiguous public/private boundaries; \\ bystander exposure; crowd navigation recording\end{tabular} \\
\rowcolor[HTML]{EFEFEF} 
\textbf{\begin{tabular}[c]{@{}l@{}}\textit{Accessibility and}\\ \textit{bystander contexts}: \\ mobile service robots \\ and drones\end{tabular}} & \begin{tabular}[c]{@{}l@{}}Sidewalks, streets, \\ delivery routes, campuses, and \\ mixed public/private spaces\end{tabular} & \begin{tabular}[c]{@{}l@{}}General public, \\ people with visual \\ impairments,\\ nearby residents, \\ operators\end{tabular} & \begin{tabular}[c]{@{}l@{}}Privacy inequity for people \\ who cannot perceive sensors; \\ unclear recording status; \\ accessibility–privacy trade-offs; \\ wide-area sensing\end{tabular} \\ \bottomrule
\end{tabular}
\end{table*}

\section{Design Implications: Toward Embodied Privacy Mechanisms}
The capability × context matrix suggests that robot privacy should be addressed through the robot’s embodied behavior, not only through data settings. The implications below follow from the boundary mechanisms identified in the framework: mobility and viewpoint control motivate spatial and sensing constraints; remote operation motivates presence disclosure and local override; manipulation motivates object- and body-level access rules; and social expression motivates attention to privacy influence. Context further determines whose boundary claims should be recognized and negotiated. We highlight six design implications.

\textbf{Treat spatial entry as a privacy event.}
Robots should treat entering, approaching, and looking into a space as privacy-relevant actions. Doorways, bedsides, bathrooms, desks, and private rooms can serve as boundary checkpoints where the robot slows down, stops, states its task, and waits for permission before entering or orienting its sensors. Designers can implement these checkpoints through navigation constraints, task disclosure, permission prompts, and fallback behaviors. Policymakers and organizations can define default no-entry or permission-required zones in sensitive settings. Users and affected occupants should be able to mark local boundaries through simple mechanisms, such as room labels, temporary no-go zones, verbal commands, or physical markers. When permission is absent, the robot should default to lower-access alternatives, such as waiting outside, leaving an item at the threshold, or sending a non-visual notification.

\textbf{Viewpoint-aware sensing control.}
For mobile robots, privacy depends on where the robot is, what direction it faces, and what its sensors can see. Privacy controls should therefore include viewpoint and field-of-view constraints. For example, robots can avoid pointing cameras at screens, documents, beds, or bathrooms; use downward-facing navigation when detailed vision is unnecessary; lower sensing resolution in sensitive areas; and make the current sensing direction visible through gaze, lights, or body orientation.

\textbf{Remote-presence disclosure and local override.}
Remote operation should be visible to people near the robot. When a remote operator can see, hear, move, or act through the robot, the robot should enter a distinct remote-presence mode. This mode can include persistent local indicators, operator identification when appropriate, and local controls to pause video, freeze movement, block entry into private rooms, or return the robot to a neutral location.

\textbf{Specify object- and body-level access rules.}
Manipulation capabilities require privacy rules beyond room-level access. Opening drawers, moving documents, handling clothing, touching bodies, or retrieving medication should be governed as privacy-relevant actions. Robots can distinguish between public, personal, sensitive, and restricted objects, and between no-contact assistance, peripheral contact, intimate care, and emergency intervention. Each category should have different defaults for permission, logging, supervision, and fallback behavior.

\textbf{Account for social expression as privacy influence}
Robots should treat social expression as privacy-relevant capability. Human-like gestures, pet-like behavior, affective responses, deference, companionship, and care-oriented interaction can change how people interpret the robot’s presence, how much they disclose, and whether they feel able to refuse. Designers should therefore evaluate not only what the robot senses or accesses, but also how its social role and expressive behavior shape privacy judgment. Privacy-sensitive designs may require limiting emotionally persuasive cues during data collection, separating companionship from monitoring functions, making institutional or commercial roles explicit, and avoiding expressions that obscure remote operators, organizational interests, or data use.

\textbf{Support local boundary claims in shared spaces.}
Robots should account for multi-user and bystander affected by their actions, not only those who own or configure them. Cohabitants, children, visitors, patients, workers, and bystanders may each hold legitimate boundary claims over rooms, objects, bodies, or situations. When such claims conflict with the preferences of an owner, caregiver, employer, or service provider, robots should default to lower-access behavior, such as stopping, stepping back, looking away, leaving the space, or requesting human mediation. To protect interruption rights, robots should also provide simple ``stop'' mechanisms that require no prior setup or technical expertise, such as visible physical buttons, direct verbal commands, gestures, or environmental markers that allow affected people to immediately halt, redirect, or distance the robot.

These mechanisms also involve trade-offs: restricting sensing, movement, or access may reduce task performance, accessibility, or safety, requiring privacy protections to be calibrated to the context and task.

\section{Research agenda and conclusion}
The capability × context view raises a research agenda for studying robot privacy as an embodied boundary problem. Future work should examine how privacy boundaries are perceived, crossed, negotiated, and repaired when robots move through shared spaces, change viewpoints, manipulate objects, mediate remote presence, and act as social partners.

First, how should robots identify privacy-relevant boundaries in everyday environments? Spatial thresholds such as bedrooms, bathrooms, desks, bedsides, and care areas may require different entry rules, but these boundaries are often socially defined rather than physically explicit. Second, how can robots make viewpoint, sensing direction, and remote presence legible to nearby people, including bystanders and people with limited ability to perceive sensors? Third, what permission models are needed for object handling and bodily assistance, where privacy concerns involve touching, opening, retrieving, or rearranging rather than only recording? Fourth, how do social expressions, roles, and relational framings affect privacy judgment, disclosure, vigilance, and refusal? Robots may cross privacy boundaries not only by entering spaces or collecting data, but also by appearing caring, harmless, deferential, companion-like, or pet-like in ways that reshape users’ expectations and responses. Fifth, how should robots handle conflicting boundary claims among owners, cohabitants, patients, caregivers, workers, visitors, and bystanders? Sixth, how can interruption rights be made available at the moment of exposure, without requiring prior setup, ownership, or technical expertise?

These questions call for more prospective work on the concrete pathways through which robot capabilities produce privacy violations. Rather than treating privacy risk as a general concern attached to robots, future research should specify how particular capabilities enable particular forms of boundary crossing: how mobility creates spatial access, how viewpoint control creates observational access, how manipulation creates object and bodily access, how teleoperation transfers local presence to remote actors, and how social expression affects disclosure, trust, and refusal. We argue that privacy-preserving robot design should therefore address both physical and social forms of embodied access. Robots should not only minimize data collection; they should minimize, disclose, and negotiate the ways they enter spaces, approach bodies, access objects, mediate remote presence, and shape social relationships.
Future work can operationalize these questions through scenario-based experiments, Wizard-of-Oz studies, multi-stakeholder interviews, and in-situ deployments that examine how people perceive, negotiate, and respond to embodied boundary crossings.


\balance
\bibliographystyle{ACM-Reference-Format}
\bibliography{0-reference}

\appendix


\end{document}